\documentclass[11pt]{article}
\usepackage{acl2016}
\usepackage{times}
\usepackage{latexsym}
\usepackage{url}
\usepackage{graphicx}
\usepackage[utf8]{inputenc}
\usepackage[english]{babel}
\usepackage{multirow}
\usepackage{float}
\usepackage{subfig}
\usepackage[export]{adjustbox}

\aclfinalcopy
\def\aclpaperid{31} %  Enter the acl Paper ID here
\title{Improved Parsing for Argument-Clusters Coordination}

\author{Jessica Ficler \\
 Computer Science Department \\
 Bar-Ilan University \\
 Israel \\
 {\tt jessica.ficler@gmail.com} \\\And
 Yoav Goldberg \\
 Computer Science Department \\
 Bar-Ilan University \\
 Israel \\
 {\tt yoav.goldbreg@gmail.com} \\}

\date{}

\begin{document}
\maketitle
\begin{abstract}
    Syntactic parsers perform poorly in prediction of Argument-Cluster
    Coordination (ACC).
    We change the PTB representation of ACC to be more suitable for learning by a
    statistical PCFG parser, affecting 125 trees in the training set.
    Training on the modified trees yields a slight improvement in EVALB scores
    on sections 22 and 23. The main evaluation is on a corpus of 4th
    grade science exams, in which ACC structures are prevalent.  On this corpus,
    we obtain an impressive $\times$2.7 improvement in recovering ACC
    structures compared to a parser trained on the original PTB trees.
\end{abstract}

\section{Introduction}

Many natural language processing systems make use of syntactic representations
of sentences.
These representations are produced by parsers, which often produce incorrect
analyses. Many of the mistakes are in coordination structures, and structures involving non-constituent coordination, such as Argument Cluster Coordination, Right Node-Raising and Gapping ~\cite{dowty1988type}, are especially hard.

Coordination is a common syntactic phenomena and work has been done to improve
coordination structures predication in the general case
~\cite{hogan2007coordinate,hara2009coordinate,shimbo2007discriminative,okuma2009bypassed}.
In this work we focus on one particular coordination structure: Argument Cluster
Coordination (ACC). While ACC are not common in the Penn TreeBank ~\cite{ptb},
they commonly appear in other corpora. For example, in a dataset of
questions from the Regents 4th grade science exam (the Aristo
Challenge), 14\% of the
sentences include ACC.

ACC is characterized by non-constituent sequences that are parallel in
structure. For instance, in \textit{``I bought John a microphone on Monday and
Richie a guitar on Saturday"}, the conjunction is between \textit{``John a
microphone on Monday"} and \textit{``Richie a guitar on Saturday"} which are
both non-constituents and include parallel arguments: the NPs \textit{``John"} and \textit{``Richie"}; the NPs \textit{``a microphone"} and \textit{``a guitar"}; and the PPs  \textit{``on Monday"} and \textit{``on Saturday"}.

Previous NLP research on the Argument Clusters Coordination \cite{mouret2006phrase} as well as the Penn
TreeBank annotation guidelines ~\cite{ptb,bies1995bracketing} focused
mainly on providing representation schemes capable of expressing the linguistic
nuances that may appear in such coordinations. The
resulting representations are relatively complex, and are not easily learnable
by current day parsers, including parsers that refine the grammar by
learning latent annotations \cite{petrov2006learning}, which are thought to be more agnostic to the
annotations scheme of the trees.
In this work, we suggest an alternative, simpler representation scheme which
is capable of representing most of the Argument Cluster coordination cases in the
Penn Treebank, and is better suited for training a parser. 
We show that by changing the annotation of 125 trees, we get a parser which is substantially better at handling ACC structures,
and is also marginally better at parsing general sentences.

\section{Arguments Cluster Coordination in the Penn Tree Bank}

Argument Cluster Coordinations are represented in the PTB with two or more
conjoined VPs, where the first VP contains a verb and indexed arguments, and the
rest of the VPs lack a verb and include arguments with indices corresponding to
those of the first conjoined VP. For example, consider the PTB representation of \textit{``The Q ratio was only 65\% in 1987 and 68.9\% in 1988"}:
\begin{center}
\includegraphics[scale=0.42]{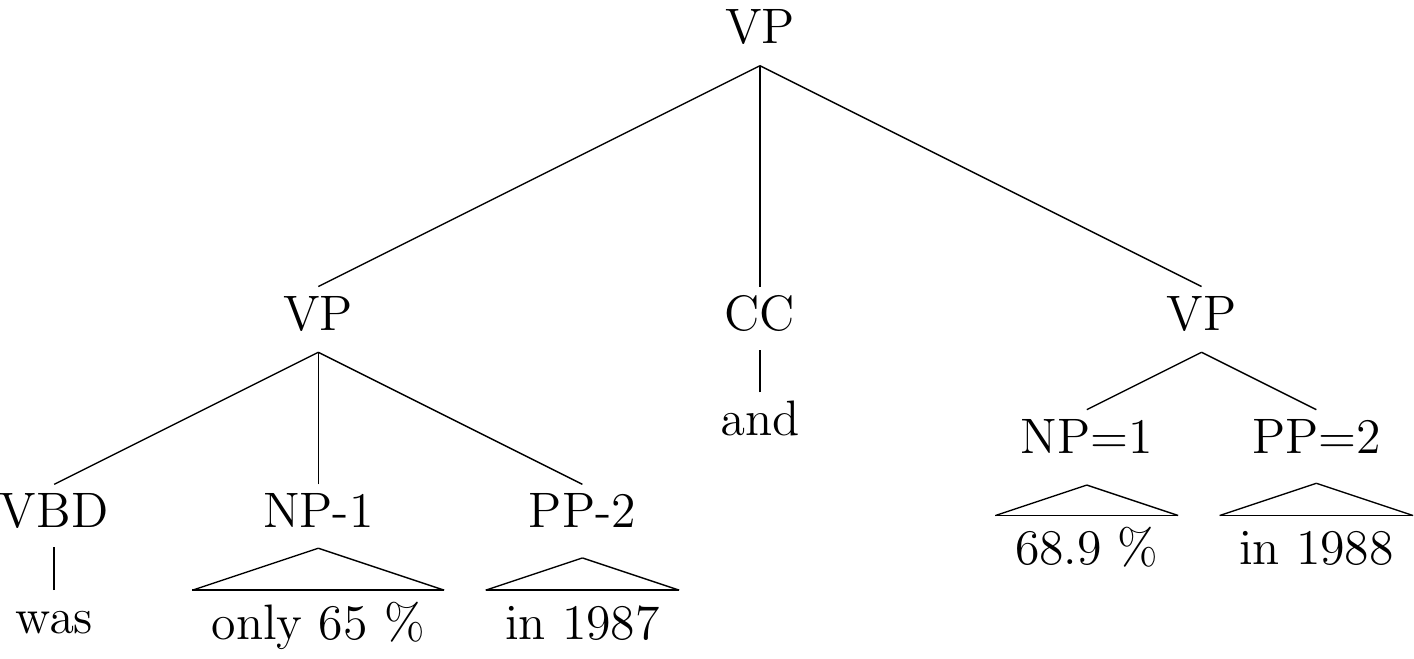} 
\end{center}
The main VP includes two conjoined VPs. The first VP includes the verb
\textit{was} and two indexed arguments: \textit{``only 65\%"} (1) and
\textit{``in 1987"} (2). The second VP does not include a verb, but
only two arguments, that are co-indexed with the parallel argument at the first
conjoined VP.

ACC structures in the PTB may include modifiers that are annotated under the
main VP,
and the conjoined VPs may includes arguments that are not part of the
cluster. These are annotated with no index, i.e. \textit{``insurance costs''} in
[\ref{fig:shared_ptb}]. 

ACC structures are not common in the PTB.  The training set includes only 141
ACC structures of which are conjoined by \textit{and} or \textit{or}. Some of them are complex but most (78\%) have the following pattern (NT is used to denote non-terminals):
\begin{center}
\includegraphics[scale=0.42]{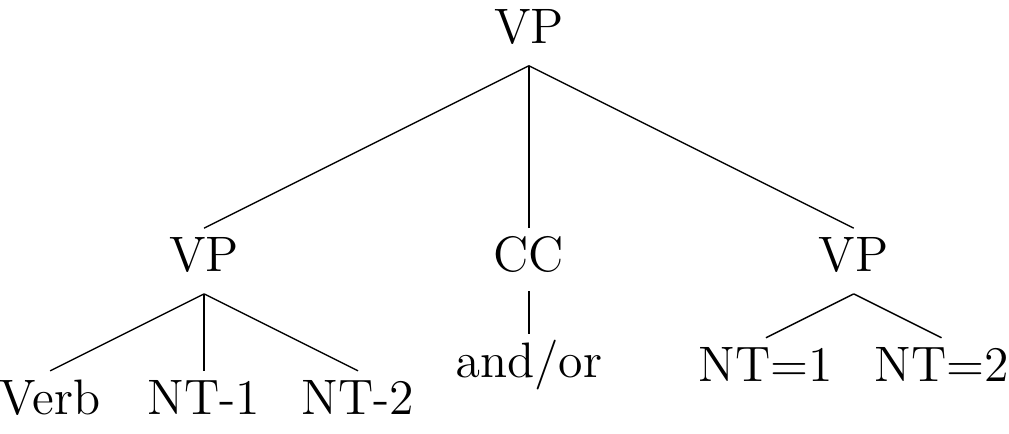}
\end{center}
These structures can be characterized as follows: (1) the first token of the first conjoined VP is a verb; (2) the indexed arguments are direct children of the conjoined VPs; (3) the number of the indexed arguments is the same for each conjoined VP.

Almost all of these cases (98\%) are symmetric: each of the conjoined VPs has the same
types of indexed arguments. Non-symmetric clusters (e.g. ``He made [these
gestures]$_{NP}^{1}$ [to the red group]$_{PP}^{2}$ and [for us]$_{PP}^{2}$
[nothing]$_{NP}^{1}$'') exist but are less common.

We argue that while the PTB representation for ACC gives a clear structure and
covers all the ACC forms, it is not a good representation for learning PCFG
parsers from. The arguments in the clusters are linked via co-indexation,
breaking the context-free assumptions that PCFG parsers rely on. PCFG parsers
ignore the indexes, essentially losing all the information about the ACC
construction.  Moreover, ignoring the indexes result in ``weird'' CFG rules such
as \texttt{VP} $\rightarrow$ \texttt{NP PP}.  Not only that the RHS of these
rules do not include a verbal component, it is also a very common structure for
NPs. This makes the parser very likely to either mis-analyze the argument
cluster as a noun-phrase, or to analyze some NPs as (supposedly ACC) VPs.
The parallel nature of the construction is also lost.
To improve the parser performance for ACC structures prediction, we suggest an alternative constituency representation for ACC phrases which is easier to learn.

\section{Alternative Representation for ACC}

Our proposed representation for ACC 
respects the context-free nature of the parser. 
In order to avoid incorrect syntactic derivations and derivations that allows conjoining of clusters with other phrases, as well as to express the symmetry that occur in many ACC phrases, we change the PTB representation for ACC as follows:
(1) we move the verb and non-indexed elements out of the first argument cluster
to under the main VP; 
(2) each argument cluster is treated as a phrase, with new non-terminal
symbols specific to argument clusters;
(3) the conjunction of clusters also receives a dedicated phrase level.
For example see comparison between the original and new representations: 
\begin{table}[H]
\begin{tabular}{lc}
\vspace{-0.5cm}
[1] & \\
\end{tabular}
\end{table}
\begin{figure}[H]
  \centering
    \subfloat[PTB representation]{\includegraphics[scale=0.39]{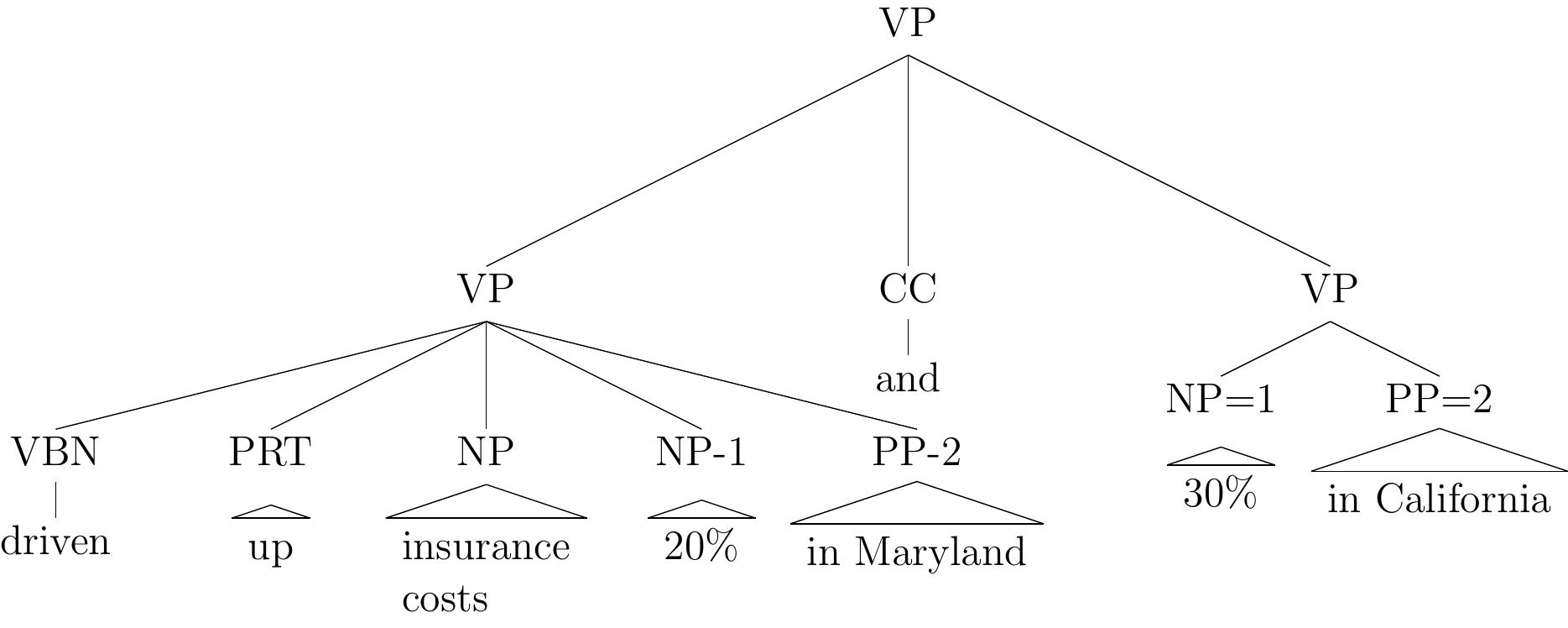}\label{fig:shared_ptb}}
    \vfill
  \subfloat[Our modified tree]
  {\includegraphics[scale=0.35]{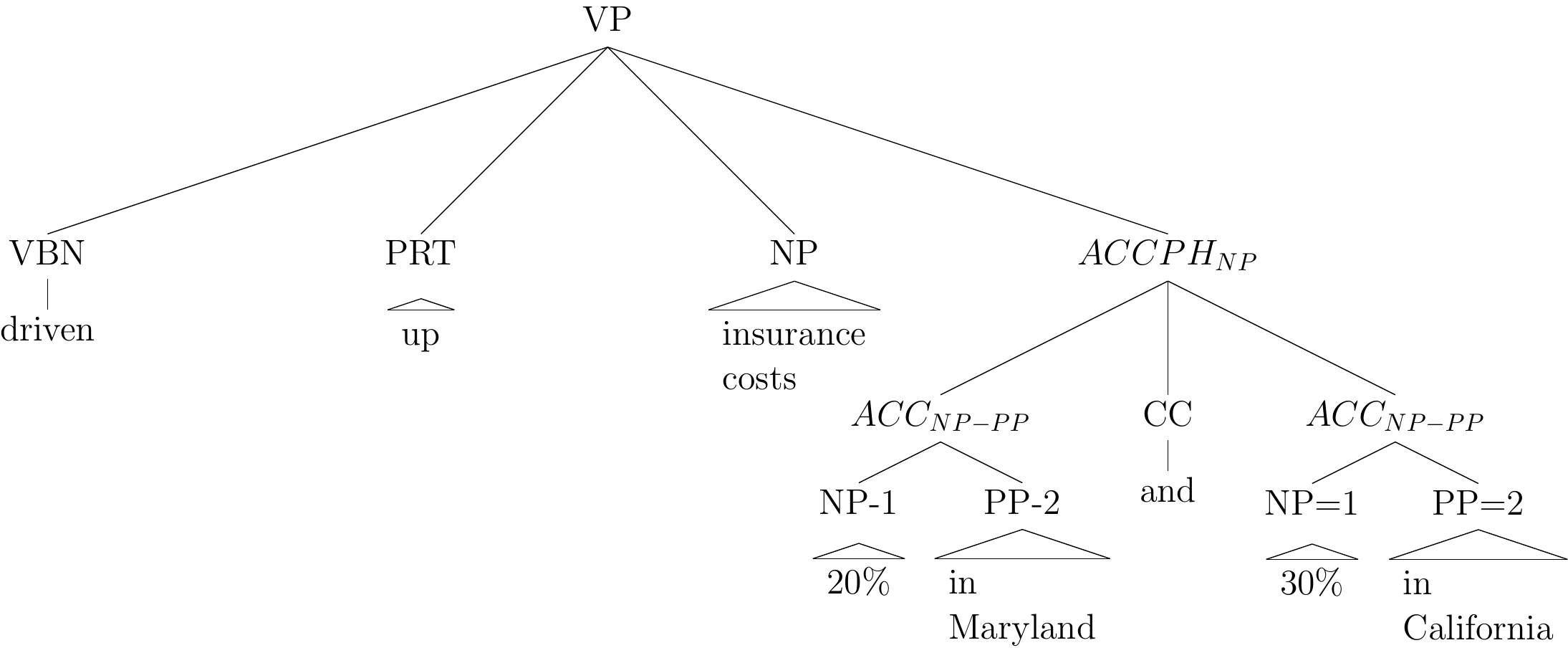}\label{fig:shared_our}}
  \label{fig:comp}
\end{figure}

The main verb \textit{driven} as well as the particle \textit{up} and the
non-indexed argument \textit{insurance costs} are moved to the external VP.
The two
argument clusters (formerly VPs) receive dedicated phrase labels $ACC_X$, where
$X$ reflects the syntactic types of the indexed elements (e.g. $ACC_{NP-PP}$ for
the first cluster in [\ref{fig:shared_our}] above).
The most common cases are $ACC_{NP-PP}$ which appears in 41.6\% of the clusters, $ACC_{ADJP-PP}$ with 21.2\% of the clusters and $ACC_{PP-PP}$ with 5.3\% of the clusters. 

Finally, we introduce a new phrase type ($ACCPH_X$) for the coordination of the two
clusters. Here $X$ denotes the main element in the clusters, determined
heuristically by taking the first of the following types that appear in any of
the clusters: NP, PP, ADJP, SBAR.
Cases where the clusters contains an ADVP element are usually special (e.g. the
following structure is missing ``people'' in the second cluster: ((NP 8000 people) (in Spain)) and ((NP 2000) (ADVP abroad))). For such cases, 
we add ``ADVP'' to the $ACCPH$ level label.
Table \ref{fig:tags_table} lists the $ACCPH$ level labels and their number of
the appearances in the 125 modified trees.\footnote{Parsers that apply latent annotations to the grammar, such as the
Berkeley Parser \cite{petrov2006learning} we use in our experiments, can potentially learn
some of our proposed refinements on their own. However, as we show in the
experiments section, the performance of the Berkeley Parser on ACC structures
significantly improve when applying our transformations prior to training.}

The representation is capable of representing common cases of ACC where the
cluster elements are siblings.
We similarly handle also some of the more complex cases, in which an extra layer
appears between an indexed argument and the conjoined VP to host an empty
element, such as in the following case with an extra S layer above
$\textit{single-B-3}$:
\begin{center}
\includegraphics[scale=0.42]{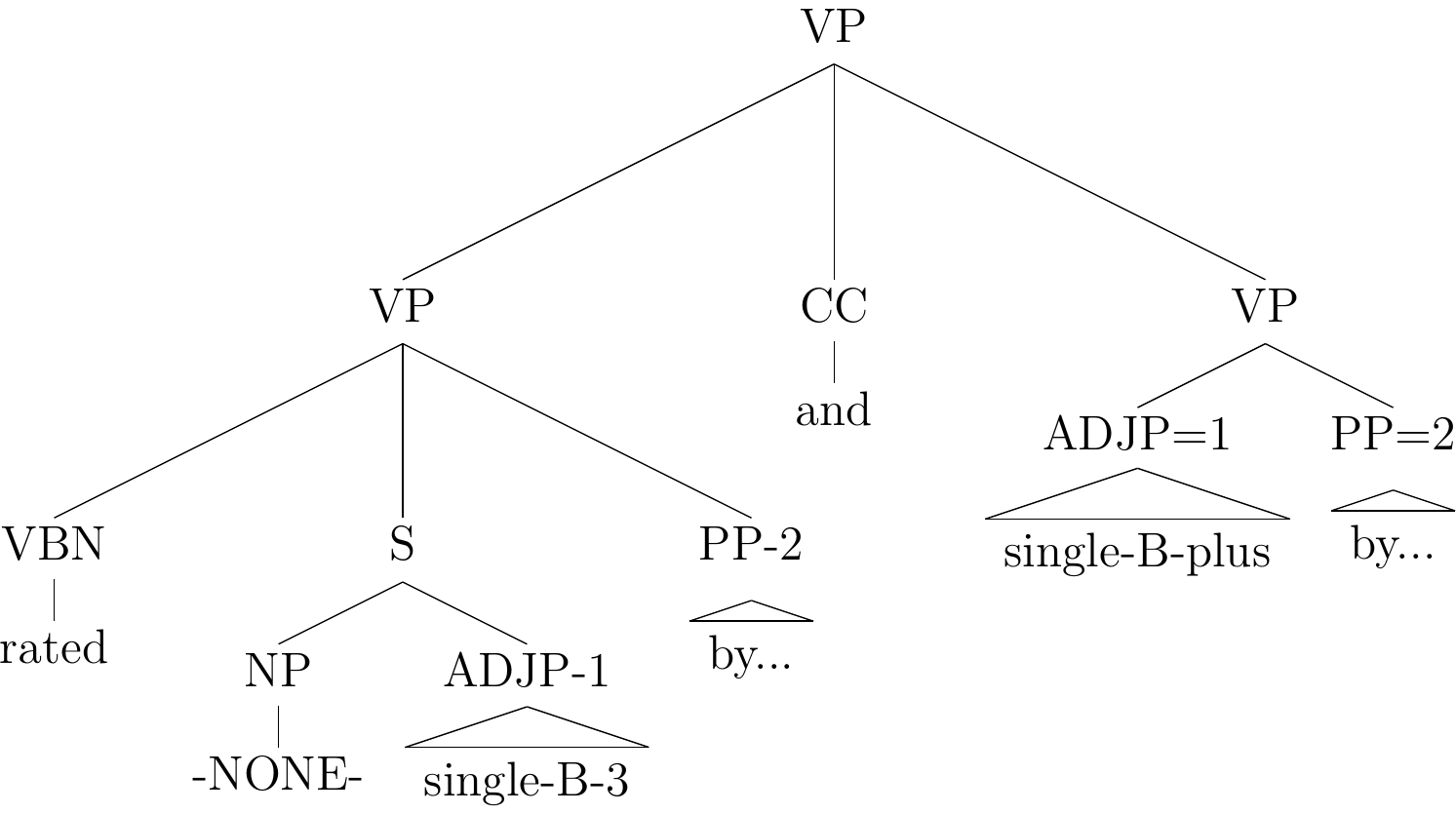}
\end{center}
in which we remove the empty NP as well as the extra S layer:
\begin{center}
\includegraphics[scale=0.42]{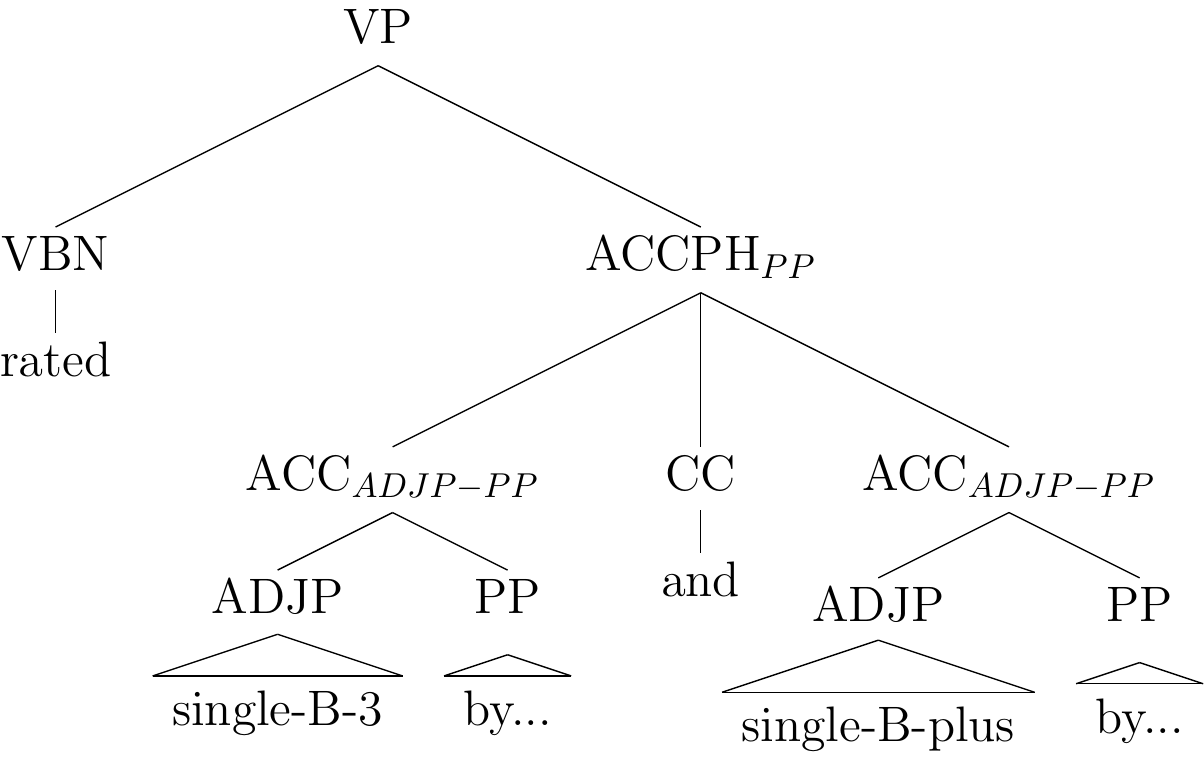}
\end{center}

\begin{table}
\begin{center}
\resizebox{\columnwidth}{!}{
\begin{tabular}{ll|ll}
\hline \bf Label &  \bf \# & \bf Label &  \bf \# \\ \hline
$ACCPH_{NP}$   & 69 & $ACCPH_{NP-ADVP}$  & 6 \\
$ACCPH_{PP}$  & 36 & $ACCPH_{PP-ADVP}$  & 11 \\
$ACCPH_{ADJP}$  & 2 & $ACCPH_{SBAR-ADVP}$  & 1 \\
\hline
\end{tabular}
}
\end{center}
\caption{The labels for the new level in the ACC trees. \#: number of occurrences. }
\label{fig:tags_table}
\end{table}

\paragraph{Limitations}
Our representation is similar to the representation that was suggested for ACC
by Huddleston et al.~\shortcite{huddleston2002cambridge} in their comprehensive
linguistic description of the English grammar.
However, while it is capable of representing the common cases of ACC, it does
not cover some complex and rare cases encountered in the PTB: 
(1) Argument-Cluster structures that include errors such as missing indexed argument and a wrong POS tag for the main verb;
(2) ACC constructions where the main verb is between the indexed arguments such
as the following: ``([About half]$_{1}$ invested [in government bonds]$_{2}$)
and ([about 10\%]$_{1}$  [in cash]$_{2}$)'';
(3) Argument-Cluster structures that include an indexed phrase which is not a direct child of the cluster head and has non-empty siblings, such as in the following case that includes an indexed argument (\textit{8\%}) which is not directly under the conjoined VP and has non-empty sibling (\textit{of}): \textit{``see a raise [[of] [8\%]$_{NP-1}$]$_{PP}$ in the first year] and [7\%]$_{NP=1}$ in each of the following two years"}.

Our changes are local and appear in small number of trees (0.003\% of the PTB train set). We also ignore more complex cases of ACC.
Yet, training the parser with the modified trees significantly improves the parser results on ACC structures.

\section{Experiments}
We converted 125 trees with ACC structures in the training sets (sections 2-21) of the PTB to the new representation, and trained the Berkeley parser \cite{petrov2006learning} with its default settings.

As the PTB test and dev sets have only 12 ACC structures that are coordinated by
\textit{and} or \textit{or}, we evaluate the parser on Regents, a dataset in
which ACC structures are prevalent (details below). As Regents does not include
syntactic structures, we focus on the ACC phenomena and evaluate the parsers’ ability to correctly identify the spans of the clusters and the arguments in them.

To verify that the new representation does not harm general parsing performance,
we also evaluate the parer on the traditional development and test sets (sections
22 and 23). As can be seen in Table \ref{tbl:ptb_results}, the parser results
are slightly better when trained with the modified trees.\footnote{The
same trend holds also if we exclude the 12 modified trees from the evaluation sets.}

\begin{table}
\begin{center}
\resizebox{\columnwidth}{!}{
\begin{tabular}{clccc}
\hline   \bf Dataset &  \bf   & \bf R & \bf P & \bf F1 \\ \hline
    \multirow{2}{*}{Dev}&PTB Trees&90.88&90.89&90.88\\
    &Modified Trees&90.97&91.21&91.09\\
\hline
    \multirow{2}{*}{Test}&PTB Trees&90.36&90.79&90.57\\
    &Modified Trees&90.62&91.06&90.84\\
    \hline
\end{tabular}
}
\end{center}
\caption{Parsing results (EVALB) on PTB Sections 22 (DEV) and 23 (TEST).}
\label{tbl:ptb_results}
\end{table}

\begin{table}
\begin{center}
\begin{tabular}{l|cc}
\hline  & \bf PTB Trees &  \bf Modified Trees \\ \hline
$ACC_{PTB}$  & 13.0 & - \\
$ACC_{OUR}$  & 24.1 & 64.8\\
\hline
\end{tabular}
\end{center}
\caption{The parser Recall score in recovering ACC conjunct spans on the Regents
dataset.   $ACC_{PTB}$: the set is annotated with the verb inside the
first cluster.  $ACC_{OUR}$: the set is annotated following our approach.}
\label{tbl:qp_results}
\end{table}
\subsection{Regents data-set}
Regents -- a dataset of questions from the Regents 4th grade science exam (the Aristo Challenge),\footnote{http://allenai.org/content/data/Regents.zip} includes 281 sentences with coordination phrases, where 54 of them include Argument Cluster coordination.
We manually annotated the sentences by marking the conjuncts spans for the
constituent coordination phrases, e.g.:

\vspace{5pt}
\noindent
\textit{Wendy (ran 19 miles) and
(walked 9 miles)} 
\noindent
\vspace{5pt}

\noindent as well as the spans of each component of the argument-cluster
coordinations, including the inner span of each argument:

\vspace{5pt}
\noindent\textit{Mary paid ([\$11.08] [for berries]) , ([\$14.33] [for apples]) , and ([\$9.31] [for
peaches])} 
\vspace{5pt}

\noindent The bracketing in this set follow our proposed ACC bracketing,
and we refer to it as $ACC_{OUR}$.  

We also created a version in which the
bracketing follow the PTB scheme, with the verb included in span of the
first cluster, e.g.:

\vspace{5pt}
\noindent
\textit{Mary
([paid] [\$11.08] [for berries]) , ([\$14.33] [for apples]) , and ([\$9.31] [for
peaches])}
\vspace{5pt}

\noindent We refer to this dataset as $ACC_{PTB}$.

We evaluate the parsers' ability to correctly recover the components of the
coordination structures by computing the percentage of gold annotated phrases 
where the number of predicted conjunct is correct and all conjuncts spans (round brackets) are predicted correctly (Recall).
For example, consider the following gold annotated phrase:

\vspace{5pt}
\noindent
\textit{A restaurant served (9 pizzas during lunch) and (6 during dinner) today}
\vspace{5pt}

A prediction of (\textit{``9 pizzas during lunch"}, \textit{``6 during dinner today"}) is considered as incorrect because the second conjunct boundaries are not matched to the gold annotation.

We compare the Recall score that the parser achieves when it is trained on the modified trees to the score when the parser is trained on the PTB trees.

When evaluated on all coordination cases in the Regents dataset (both ACC and
other cases of constituent coordination), the parser trained on the modified trees
was successful in recovering 54.3\% of the spans, compared to only 47\% when
trained on the original PTB trees.

We now focus on specifically on the ACC cases (Table
\ref{tbl:qp_results}).  
When evaluating the PTB-trained parser on $ACC_{PTB}$, it correctly recovers
only 13\% of the ACC boundaries.  Somewhat surprisingly, the PTB-trained
parser performs \emph{better} when evaluated against $ACC_{OUR}$, correctly
recovering 24.1\% of the structures.  This highlights how unnatural the original
ACC representation is for the parser: it predicts the alternative representation
more often than it predicts the one it was trained on.
When the parser is trained on the modified trees, results on $ACC_{OUR}$ jump to
64.8\%, correctly recovering $\times$2.7 more structures.

The previous results were on recovering the spans of the coordinated elements
(the round brackets in the examples above).  When measuring the Recall in
recovering any of the arguments themselves (the elements surrounded by square
brackets), the parser trained on the modified trees recovers 72.46\% of the
arguments in clusters, compared to only 58.29\% recovery by the PTB-trained parser. We also measure
in what percentage of the cases in which both the cluster boundaries (round
brackets) were recovered correctly, all the internal structure (square brackets)
was recovered correctly as well.
The score is 80\% when the parser trained on the modified trees compared to 61.5\% when it is trained on the PTB-trees.

Overall, the parser trained on the modified trees significantly outperforms the
one trained on the original trees in all the evaluation scenarios.

Another interesting evaluation is the ability of the parser that is trained on
the modified trees to determine whether a coordination is of Argument Clusters
type (that is, whether the predicted coordination spans are marked with the ACCPH label).\footnote{This measurement is relevant only when parsing based on our proposed
annotation, and cannot be measured for parse trees based the original PTB annotation.}
The results are a Recall of 57.4\% and Precision of 83.78\%.
When we further require that both the head be marked as ACCPH and the internal structure be correct, the results are 48.14\% Recall and 70.27\% Precision.

\section{Conclusions}
By focusing on the details of a single and relatively rare syntactic
construction, argument clusters coordination, we have been able to significantly improve parsing results for
this construction, while also slightly improving general parsing results.
More broadly, while most current research efforts in natural language processing and in 
syntactic parsing in particular is devoted to the design of general-purpose,
data-agnostic techniques, such methods work on the common phenomena while often
neglecting the very long tail of important constructions.
This work shows that there are gains to be had also
from focusing on the details of particular linguistic phenomena, and changing
the data such that it is easier for a ``data agnostic'' system to learn.

\section*{Acknowledgments}
This work was supported by The Allen Institute for Artificial Intelligence
as well as the German Research Foundation via the
German-Israeli Project Cooperation (DIP, grant DA 1600/1-1).
\bibliography{acl2016}
\bibliographystyle{acl2016}
\end{document}